%% file: main.tex
\pdfoutput=1
\documentclass[11pt]{article}

\usepackage[preprint]{acl}

\usepackage{times}
\usepackage{latexsym}
\usepackage[T1]{fontenc}
\usepackage[utf8]{inputenc}
\usepackage{microtype}
\usepackage{inconsolata}
\usepackage{graphicx}
\usepackage{booktabs}
\usepackage{multirow}
\usepackage{amsmath}
\usepackage{amssymb}
\usepackage{xcolor}
\usepackage{pifont}
\usepackage{enumitem}
\usepackage{caption}
\setlist{nosep,leftmargin=*}
\newcommand{\bench}{\textsc{HWRE-Bench}}
\newcommand{\cmark}{\ding{51}}
\newcommand{\xmark}{\ding{55}}
\newcommand{\pmark}{$\circ$}
\newcommand{\finding}[1]{\paragraph{#1}}

\title{\bench: Can LLM Agents Understand Hardware They Did Not Design?}

\author{
  \textbf{Peilong Zhou}\textsuperscript{1,2,3},
  \textbf{Cangyuan Li}\textsuperscript{1,2},
  \textbf{Kaiyan Chang}\textsuperscript{1,2},
  \textbf{Ying Wang}\textsuperscript{1,2}
  \\
  \textsuperscript{1}SKLP, Institute of Computing Technology, Chinese Academy of Sciences \\
  \textsuperscript{2}University of Chinese Academy of Sciences \quad
  \textsuperscript{3}School of Advanced Interdisciplinary Sciences \\
  \small{\textbf{Correspondence:} \href{mailto:wangying2009@ict.ac.cn}{wangying2009@ict.ac.cn}}
}

\begin{document}
\maketitle

\begin{abstract}
Hardware benchmarks for LLM agents all run the design flow forward, from a
specification to RTL, a layout or a bug fix. We ask the reverse question: given
an implemented artifact, such as a layout, a gate-level netlist or a protected
simulation model, can an agent work out what it does well enough to act on that
knowledge? We introduce \bench{}, a benchmark for agentic hardware reverse
engineering built on one observation. Each step of the forward flow discards
information (names, hierarchy, word-level structure, generator parameters, the
programming model), and a reverse-engineering task asks the agent to recover
what was lost and to prove it with an executable deliverable that the artifact
alone cannot supply. Since layout extraction is now mechanical and a correct
netlist is already a cycle-exact model, every task carries a mandatory
\emph{copy-the-netlist} control that must score zero. \bench{} spans five tracks
(Extract, Abstract, Audit, Interoperate, Modify), four information-loss levels
and white- versus black-box access, all on original designs. We report a pilot
of 148 rollouts from two agents on four seed tasks under 16 conditions, with stage-annotated
trajectories. Agents built their own layout extractors and audited a sanitized
layout as reliably as a netlist, at about 27\% more turns. Keyless,
equivalence-preserving obfuscation changed neither pass rates nor
defect-repair scores. Whether agents read the structure at all depended on task
framing rather than on necessity, so pass rate alone does not show
understanding. Provider safety filters stopped 16\% of one agent's runs on the
audit tasks. These results argue for deliverables beyond
equivalence and for reporting run validity explicitly.
\end{abstract}

\begin{figure*}[t]
\centering
\includegraphics[width=\textwidth]{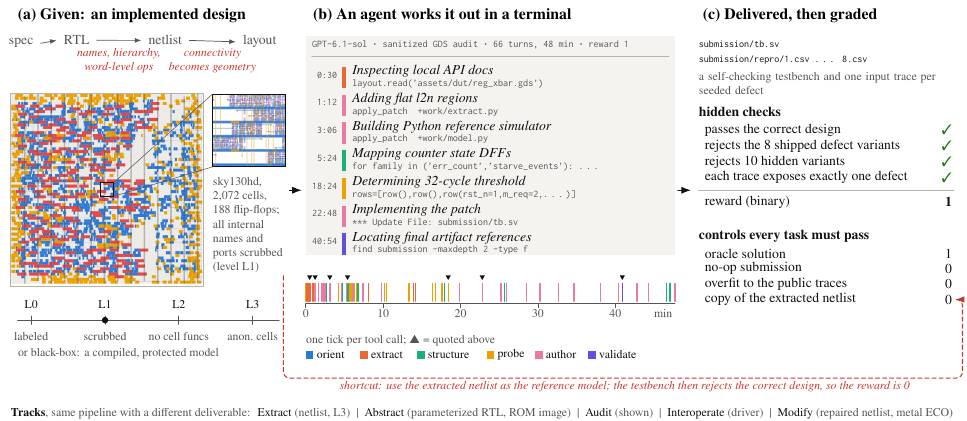}
\caption{\bench{} on one real run. \textbf{(a)} The agent receives an
implemented design, here the sanitized sky130 layout of the original xbar
design at information-loss level L1; the gray chain above it shows what the
forward flow threw away. \textbf{(b)} An excerpt of a passing GPT-6.1-sol run
on the layout audit task: the agent's own step headings (italic) and the
commands it ran (abridged), each colored by the stage our annotation rules
assign (Section~\ref{sec:setup}); the barcode below shows every tool call of
the 48-minute run. \textbf{(c)} The deliverable is graded on hidden checks, and
every task must pass four controls. The red path is the transcription shortcut
that the controls close: reusing the extracted netlist gives a reward of 0.}
\label{fig:overview}
\end{figure*}

\section{Introduction}

Large language model (LLM) agents now write, verify and optimize RTL
\citep{liu2023verilogeval,lu2024rtllm,pinckney2025cvdp}, and they operate
terminal toolchains for long horizons \citep{jimenez2024swebench,tbench2025}.
In software, the same capability has been pointed backward: a growing set of
benchmarks asks agents to reverse-engineer binaries, solve crackmes and capture
flags \citep{shao2024nyuctf,zhang2024cybench,crackmebench2026,programbench2026}.
Hardware has no counterpart. Existing hardware agent benchmarks all go forward
\citep{pinckney2025cvdp,hwebench2026,pdagentbench2026,closerbench2026}, and
prior work on hardware reverse engineering (HWRE) evaluates single-step models
or human-in-the-loop tools, such as gate-level function identification
\citep{alrahis2021gnnre,wu2023gamora,fang2025geneda} or interactive netlist
analysis \citep{fyrbiak2019hal}. A recent systematization of two decades of
HWRE research finds that reproducible artifacts and benchmarks are scarce
\citep{sok2026silicon}.

The gap matters for two reasons. First, understanding hardware one did not
design is routine engineering: supply-chain and trust audits, drivers for
undocumented peripherals, second sourcing of discontinued parts, post-silicon
errata, and metal-only engineering change orders. Forward benchmarks measure
whether an agent can \emph{design}; they say nothing about whether it can
\emph{read and change} an implementation written by someone else. Second, the
cost of reverse engineering is what an IP owner must estimate before deciding
how much to protect, and anecdotal evidence suggests that the cost is collapsing.
In one of our case studies (Section~\ref{sec:softmax}), an agent recovered a
cycle-exact RTL model of a sanitized 7\,nm softmax accelerator from its GDSII
layout alone in under ten minutes.

Building such a benchmark is not a matter of packaging layouts. Layout-to-netlist
extraction is now mechanical, and a functionally correct netlist is already a
cycle-exact model. A task that asks for ``an RTL equivalent to this chip'' can
be solved by transcribing the extracted netlist, and an equivalence checker
awards full marks. That task measures extraction, not understanding. \bench{}
is therefore built on three principles (Figure~\ref{fig:overview}):
\begin{itemize}
\item \textbf{Anti-transcription grading.} Every deliverable is something the
  artifact cannot provide: a design at new generator parameters, a new
  microcode image, a driver, distinguishing stimuli, or a repaired netlist.
  Each task ships with a copy-the-netlist control that must score 0.
\item \textbf{Hidden but fair semantics.} Difficulty sits in behavior that
  white-box reading can derive but black-box probing rarely hits (counter
  widths, zero-means-maximum encodings, saturation versus wrap-around, clear
  conditions). Hidden tests probe only derivable behavior.
\item \textbf{Original, generated designs.} All designs are original and come
  from parameterized families, so fresh instances can be generated at chosen
  scales and information-loss levels.
\end{itemize}

We validate the design with a pilot of 148 rollouts from two agents on four
seed tasks under 16 conditions. We annotate every tool call with a reverse-engineering
stage, which turns trajectories into evidence about \emph{how} the agents work
and not only whether they pass. Our main findings are:
\begin{enumerate}
\item Layout delivery is not the bottleneck. Agents wrote their own KLayout
  extractors, found and fixed their own extraction bugs, and audited a sanitized
  layout as reliably as a netlist, at about 27\% more turns
  (Section~\ref{sec:gds}).
\item Pass rate does not reveal understanding. On the same design, a change in
  task wording moved the share of structural analysis from 33\% to 2\% of tool
  calls with no change in pass rate (Section~\ref{sec:how}).
\item Keyless, equivalence-preserving obfuscation did not slow agents. Pass
  rates, defect-repair scores and netlist reads were unchanged
  (Section~\ref{sec:obf}).
\item Run validity must be reported. Provider safety filters stopped 7 of 44
  runs (16\%) of one agent on the audit tasks, and agent, API and sandbox
  failures invalidated another 27\% of all slots (Section~\ref{sec:validity}).
\end{enumerate}

\paragraph{Contributions.} (1) A formulation of agentic HWRE as the recovery of
information lost by the forward flow, with five tracks, four information-loss
levels and two access modes (Section~\ref{sec:bench}). (2) An anti-transcription
grading protocol with mandatory negative controls (Section~\ref{sec:grading}).
(3) An open build pipeline from original RTL through OpenROAD to sanitized
layouts with an extraction check (Section~\ref{sec:pipeline}). (4) A pilot
study with stage-annotated trajectories, run-validity accounting and
statistical tests (Sections~\ref{sec:setup}--\ref{sec:results}).

\section{Related Work}
\label{sec:related}

\begin{table}[t]
\centering
\small
\setlength{\tabcolsep}{2.2pt}
\resizebox{\columnwidth}{!}{%
\begin{tabular}{lcccccc}
\toprule
 & HW & Agent & Rev. & Layout & Exec.\ & Anti- \\
 &  &  &  & input & deliv. & copy \\
\midrule
VerilogEval, RTLLM & \cmark & \xmark & \xmark & \xmark & \cmark & \xmark \\
CVDP, HWE-Bench & \cmark & \cmark & \xmark & \xmark & \cmark & \xmark \\
PDAgent, CLOSER & \cmark & \cmark & \xmark & \pmark & \cmark & \xmark \\
GNN-RE, Gamora & \cmark & \xmark & \cmark & \xmark & \xmark & \xmark \\
GenEDA & \cmark & \xmark & \cmark & \xmark & \pmark & \xmark \\
NYU CTF, Cybench & \xmark & \cmark & \cmark & \xmark & \cmark & \xmark \\
CrackMe-, ProgramBench & \xmark & \cmark & \cmark & \xmark & \cmark & \pmark \\
\midrule
\bench{} (ours) & \cmark & \cmark & \cmark & \cmark & \cmark & \cmark \\
\bottomrule
\end{tabular}}
\caption{Positioning. \emph{Rev.}: the artifact is the only source of truth;
\emph{Layout input}: tasks start from GDSII; \emph{Exec.\ deliv.}: graded by
executing a deliverable; \emph{Anti-copy}: a control rejects transcribing the
given artifact. \pmark{} = partially.}
\label{tab:related}
\end{table}

\paragraph{Hardware agent benchmarks.}
VerilogEval \citep{liu2023verilogeval} and RTLLM \citep{lu2024rtllm} evaluate
RTL generation from specifications. CVDP \citep{pinckney2025cvdp} adds
verification and debugging tasks; newer agent benchmarks cover real hardware
bug repair \citep{hwebench2026}, physical design \citep{pdagentbench2026} and
cross-stage design closure \citep{closerbench2026}. All of them treat the
specification or RTL as ground truth and move toward an implementation.
\bench{} reverses the direction (Table~\ref{tab:related}).

\paragraph{Software reverse engineering for agents.}
Capture-the-flag suites \citep{shao2024nyuctf,zhang2024cybench} include
reversing challenges, and dedicated benchmarks now target binary reversing
\citep{crackmebench2026}. ProgramBench \citep{programbench2026} asks agents to
rebuild programs and blocks solutions that wrap the original executable, which
is the software analogue of our copy-the-netlist control. A binary keeps
instruction-level semantics, whereas synthesis erases word-level operators and
hierarchy, so the hardware setting differs in kind rather than in degree.

\paragraph{Hardware reverse engineering.}
Classic work recovers word-level modules from flat netlists by structural and
functional analysis \citep{subramanyan2014reverse,torrance2009state,quadir2016survey}.
Graph neural networks classify gates into modules or identify arithmetic blocks
\citep{alrahis2021gnnre,wu2023gamora}, and LLM-based work maps netlists to
functional descriptions or RTL in one step \citep{fang2025geneda}. HAL
\citep{fyrbiak2019hal} supports human analysts interactively. None of these
evaluates an autonomous agent that must chain extraction, simulation, hypothesis
testing and an executable deliverable over a long horizon.

\paragraph{Hardware obfuscation.}
Logic locking and camouflaging protect designs with secrets
\citep{rajendran2015fault,yasin2017provably}; keyless structural obfuscation
only raises analysis cost. We exclude tasks that defeat protection mechanisms
(Section~\ref{sec:ethics}). Our obfuscation study uses only keyless,
equivalence-preserving transformations of our own designs, to measure whether
structure-level defenses affect agents at all.

\section{\bench{}}
\label{sec:bench}

\subsection{Reverse engineering as recovery of lost information}

\begin{figure*}[t]
\centering
\includegraphics[width=\textwidth]{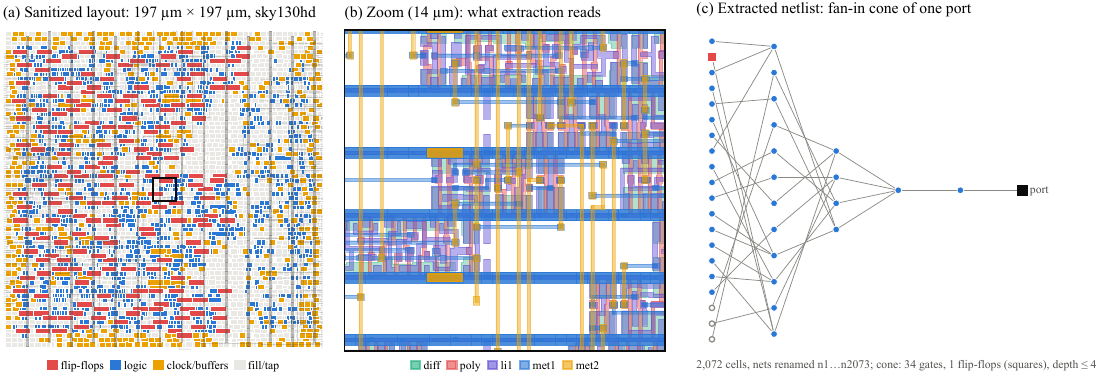}
\caption{What the agent receives for the xbar seed design (2{,}072 standard
cells, 188 flip-flops). (a) The sanitized sky130hd layout: top renamed, every
internal name and property removed, instance order shuffled; colors show cell
classes recovered from the cell library, gray strokes are met3/met4 routing.
(b) A 14\,\textmu m window of the full geometry that extraction must interpret.
(c) The fan-in cone of one output port in the netlist extracted from (a), whose
nets are anonymized (\texttt{n1}\ldots\texttt{n2073}).}
\label{fig:artifacts}
\end{figure*}

The forward flow is a chain of lossy maps (Figure~\ref{fig:overview}, top).
Synthesis drops names, hierarchy, word-level structure and generator
parameters. Place-and-route turns connectivity into geometry. Shipping a
peripheral without a register manual drops its programming model. A \bench{}
task fixes one artifact, one kind of lost information and one deliverable that
exhibits it. Figure~\ref{fig:artifacts} shows what an agent actually receives.

\subsection{Tracks}
\label{sec:tracks}

\textbf{Extract} recovers a gate-level netlist from a layout. It is the only
track where a netlist is a legitimate deliverable (graded by equivalence
checking), and it is used at the hardest information-loss level.
\textbf{Abstract} recovers structure that synthesis erased: given layouts of
three variants of one generator with their parameter manifests, deliver a
parameterized RTL graded at hidden parameter points; or, given a ROM-programmed
sequencer, deliver a new ROM image that implements a new function from
micro-operations present in the old one.
\textbf{Audit} gives a specification and an artifact with $K$ independent seeded
defects ($K$ disclosed); the agent delivers a self-checking testbench and one
reproducer per defect, graded two-sidedly (pass the correct design, reject
every single-defect mutant, each reproducer exposes exactly its own defect).
\textbf{Interoperate} gives the layout of an undocumented peripheral and its SoC
integration facts; the agent delivers a driver against a fixed C API, graded
on hidden co-simulation scenarios.
\textbf{Modify} changes the artifact itself: repair a defective netlist against
the specification, or perform a metal-only ECO with spare cells.

\subsection{Information-loss levels and access}

Each design is released at four levels. \textbf{L0}: hierarchical layout with
cell and port names. \textbf{L1}: ports renamed, all internal names and
properties removed, instance order shuffled (Figure~\ref{fig:artifacts}a).
\textbf{L2}: L1 with LEF abstracts only, without Liberty functions.
\textbf{L3}: anonymized cells whose functions must be recognized from
transistor geometry. Independently, a task is \emph{white-box} (layout or
netlist) or \emph{black-box} (a compiled, protected simulation model), which
isolates the value of structural access.

\subsection{Anti-transcription grading}
\label{sec:grading}

Every task is a container with an isolated verifier that writes a binary reward.
Before a task is admitted it must pass four controls: the oracle solution scores
1, a no-op submission scores 0, a \emph{copy-the-netlist} submission (the
extracted netlist wrapped as the deliverable) scores 0, and a submission that
hard-codes the public tests scores 0. The authors also re-derive the reference
solution from the sanitized inputs alone, which confirms that every hidden
behavior is derivable. Besides pass@$k$, the harness logs turns, tokens,
wall-clock time, artifact reads and a stage-annotated trace.

\subsection{Build pipeline}
\label{sec:pipeline}

Designs are original parameterized RTL families. Each is run through Yosys and
OpenROAD-flow-scripts on open PDKs (sky130hd and ASAP7). A sanitizer removes
names and properties for the target level, and a check flow re-extracts the
netlist from the sanitized layout and proves it equivalent to the flow's netlist
before release. Tasks are packaged in the Harbor and Terminal-Bench format
\citep{tbench2025} and run in E2B sandboxes with KLayout, Yosys/ABC, eqy, Icarus
Verilog and Verilator installed. The target release has 8--12 design families
expanded into 60--100 instances across tracks, levels and three scale points
(about 1k, 10k and 50k cells).

\section{Pilot Setup}
\label{sec:setup}

\paragraph{Seed tasks.}
The pilot uses three original designs. \textbf{xbar} is a two-master,
three-slave register crossbar with address decoding, reservations, arbitration
budgets and error counters (2{,}072 cells, 188 flip-flops). Its \emph{Audit}
task seeds 8 independent defects; we ran it from the gate-level netlist (task
revisions 2--3 and 4, which differ in wording, Section~\ref{sec:how}) and from
the sanitized L1 layout with the sky130 PDK. Its \emph{Modify} task gives the
defective netlist and asks for a repaired one, scored 0--8 by replaying eight
single-defect isolation traces against the golden design. \textbf{pmover} is a
DMA peripheral (APB slave, DMA master, level interrupt) whose register-level
behavior is undocumented; its \emph{Interoperate} task gives the L1 layout and
asks for a C driver graded on 30 hidden co-simulation scenarios. \textbf{pkr} is
a packet rewriter released only as a protected Verilator model with 8 seeded
bugs; it serves as a black-box reference. All tasks passed the oracle and no-op
controls; the copy-the-netlist control is trivially 0 for these
deliverables (a testbench, a driver, or a repaired netlist judged against
defects the given netlist contains).

\paragraph{Obfuscation conditions.}
For xbar we built three keyless variants of the delivered netlist, each proven
equivalent to it by sequential equivalence checking (ABC \texttt{dsec}):
\textbf{V0r}, the original netlist re-exported through the same AIG-to-gate
mapping as the variants (+0.4\% gates), which also flattens its word-level
vector declarations into scalars (the controlled baseline);
\textbf{T2$\times$3}, three rounds of randomized ABC resynthesis recipes
(+1.0\% AIG nodes); and \textbf{W25}, windowed selective hiding that removes the
named intermediate signals of chosen register windows at up to +25\% area.
\textbf{V0} denotes the unmodified netlist.

\paragraph{Agents.}
GPT-6.1-sol (reasoning effort xhigh) in the Codex CLI harness, and
DeepSeek-flash (effort max, the highest the harness offers) in the DeepSeek
harness (dsh). Each slot is an independent run in a fresh sandbox with a
wall-clock limit set by the task.

\paragraph{Validity.}
A run is \emph{valid} if the verifier produced a reward and the trial raised no
exception. Runs stopped by a provider safety filter, an agent or API crash, or a
sandbox failure are invalid; they are reported (Table~\ref{tab:validity}) but
never counted as task failures. In one condition the verifier sandbox failed
after the agent had finished; those submissions were re-graded locally with the
same grader.

\paragraph{Stage annotation.}
We assign every tool call to one of six stages with deterministic rules on the
command text (Appendix~\ref{app:stages}): \emph{orient} (reading the
specification and files), \emph{extract} (layout processing),
\emph{structural analysis} (reading or querying the netlist, tracing cones),
\emph{behavioral probing} (simulating, replaying stimuli, testing hypotheses
against a model), \emph{author} (writing tools, reference models or the
deliverable) and \emph{self-validate} (running the public checker). Process
polling is excluded. An \emph{artifact read} is any call whose arguments name
the DUT netlist or layout.

\paragraph{Statistics.}
Pass rates carry Wilson 95\% intervals. We compare pass counts with two-sided
Fisher exact tests and turn counts or repair scores with exact two-sided
Mann--Whitney tests. With 4--10 valid runs per cell, the pilot can detect only
large effects; we report the tests to bound what the data can and cannot say.

\section{Results}
\label{sec:results}

\input{tables/main_results}

Table~\ref{tab:main} lists every condition. Across tasks, valid-run pass rates
range from 0.00 to 0.80, median runs last 21--194 minutes and 41--270 turns,
and no task is saturated. We organize the results by four questions.

\subsection{Is a layout harder than a netlist?}
\label{sec:gds}

\begin{figure}[t]
\centering
\includegraphics[width=\columnwidth]{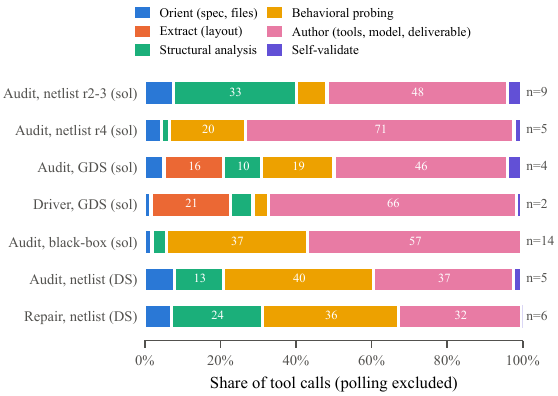}
\caption{Share of tool calls per reverse-engineering stage, pooled over valid
runs (numbers are percentages $\geq 9$). sol = GPT-6.1-sol, DS = DeepSeek-flash.
On the same xbar design, task revision 4 moved sol from structural analysis
(33\%) to authoring a model from the specification (71\%).}
\label{fig:stages}
\end{figure}

\begin{figure*}[t]
\centering
\includegraphics[width=\textwidth]{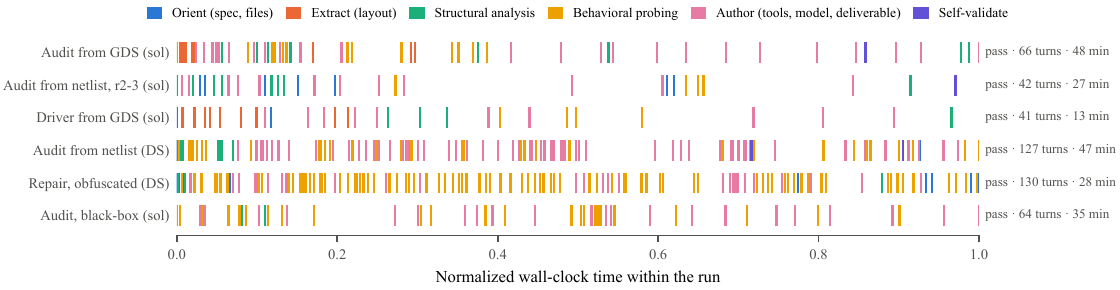}
\caption{Trajectory barcodes: one representative passing run per setting (the
run whose turn count is closest to the condition median), each tick a tool call
colored by stage and placed on normalized wall-clock time. Layout runs front-load
extraction; DeepSeek interleaves behavioral probing throughout; the black-box
run has nothing to extract or trace.}
\label{fig:barcodes}
\end{figure*}

On the xbar audit, the sanitized layout produced 2 passes in 4 valid runs
against 3 in 5 from the netlist (Fisher $p{=}1.0$), with median turns rising
from 49 to 62.5 (Mann--Whitney $p{=}0.024$) and median time from 25 to 37
minutes. The trajectories show why the cost is modest. Without being asked, the
agent loaded the GDSII with the KLayout Python API, wrote an extractor that
traces pins, metal and vias into a netlist, wrapped the sky130 Liberty cells in
a Yosys flow, and only then switched to the same analysis it uses on a netlist.
Extraction took 16\% of tool calls (Figure~\ref{fig:stages}) and was
concentrated at the start of the run (Figure~\ref{fig:barcodes}, top row). In
the softmax case study the agent also noticed a missing metal layer in its own
extraction from undriven nets and repaired it. These observations support
starting a threat model at the netlist, and placing benchmark difficulty in
semantics rather than in the input format.

\subsection{How do agents reverse-engineer?}
\label{sec:how}

Figure~\ref{fig:stages} shows four distinct strategies. With layout input, sol
extracts first and spends the rest of the run authoring analysis tools (cone
tracers, replay harnesses), a reference model and the deliverable. On the
pmover driver task, extraction took 21\% of calls and authoring 66\%: the
deliverable is software, so most of the work is iterating on the driver.
DeepSeek spends a third of its calls on behavioral probing, running many small
simulations, and on the repair task 24\% on structural analysis, where it must
localize faults in the netlist. With only a compiled model (pkr), structural
analysis disappears and probing rises to 37\%.

\finding{Pass rate does not reveal whether the structure was understood.}
xbar audit revisions 2--3 and revision 4 use the same design, the same eight
defects and the same deliverables. Revision 4 condenses the instruction from
1{,}980 to 278 words (the specification moves to a separate file), calls the
DUT a black box observed only through its ports, and asks explicitly for an
executable reference model in the testbench. Under revisions 2--3, 33\% of sol's calls analyzed
the netlist (a median of 16 artifact reads per run). Under revision 4 this fell
to 2\% (5 reads), and the agent instead wrote a reference model from the
specification and compared it with the netlist in simulation. Pass rates were
indistinguishable (6/9 against 3/5). A benchmark that reports only pass rate
cannot tell these strategies apart, and an audit task with a complete
specification can be solved largely without reading the structure. This is why
\bench{} favors deliverables that require structural knowledge, such as new
parameter points or new microcode (Section~\ref{sec:tracks}).

\subsection{Does keyless obfuscation slow agents?}
\label{sec:obf}

\begin{figure*}[t]
\centering
\includegraphics[width=\textwidth]{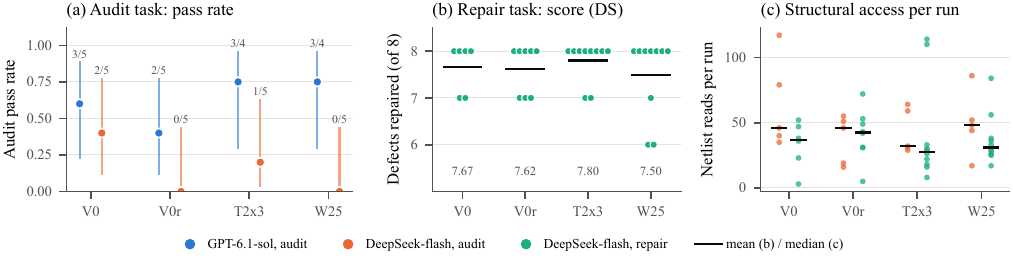}
\caption{Keyless obfuscation of the xbar netlist. (a) Audit pass rate with
Wilson 95\% intervals and pass/valid counts; the sol V0 point comes from an
earlier batch, so V0r is the controlled baseline. (b) Defects repaired out of 8
in the repair task, one dot per valid run, bar = mean. (c) Netlist reads per
run, bar = median. No contrast with V0/V0r is significant (Fisher $p\geq0.44$
for pass counts; Mann--Whitney $p\geq0.65$ for repair scores).}
\label{fig:obf}
\end{figure*}

Figure~\ref{fig:obf} compares V0/V0r with two very different keyless
transformations. For sol, T2$\times$3 and W25 each passed 3/4 against 2/5 for
V0r (Fisher $p{=}0.52$ each). For DeepSeek, pass rates were low in every
condition (0--2 of 5). On the repair task, scores sat at the ceiling in all four
conditions (means 7.50--7.80 of 8, Mann--Whitney $p\geq0.65$ against the pooled
baselines), and the number of netlist reads per run did not change in a
consistent direction. A complementary offline probe found that, after
normalization with ABC (\texttt{strash; fraig; resyn2; dch}) and simulation
against the original, nearly all internal register bits of the transformed
netlists still map one-to-one to named RTL signals, so an agent that simulates
loses little. We read this as evidence that structure-level, keyless
obfuscation is largely irrelevant to simulating agents when a specification is
available, not as proof that no obfuscation can help: the repair scores show a
ceiling effect, and a variant of the repair task with a structure-fidelity
requirement (the submission must preserve most of the given gates) is the
natural next test.

\subsection{Where do agents fail, and at what cost?}
\label{sec:fail}

\paragraph{Failures concentrate on hidden semantics.}
Of the 29 failed valid audit runs on xbar (both agents), 26 missed at least one
of the ten hidden mutants, 12 failed the reproducer-isolation check and 12
missed a shipped defect variant, while only 4 rejected the correct design. The
testbenches were usually sound but incomplete. On the repair task, all ten
imperfect runs missed one or both of the same two defects (each missed in 6
runs), and no other defect was ever missed. This is the ``hidden but fair'' regime the benchmark targets: these semantics are
in the specification and the structure, but random probing rarely reaches them.

\begin{figure}[t]
\centering
\includegraphics[width=\columnwidth]{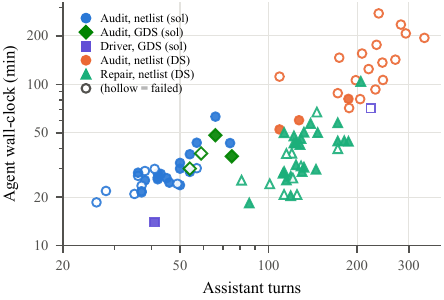}
\caption{Cost of every valid run (log--log). Filled markers passed, hollow
markers failed. The two agents occupy different regions, and the same task can
cost an order of magnitude more for one agent than for the other.}
\label{fig:cost}
\end{figure}

\paragraph{Cost and model differences.}
Figure~\ref{fig:cost} plots turns against wall-clock for every valid run.
DeepSeek used 4--5$\times$ more turns and 17--27$\times$ more input tokens
(including cached tokens) than sol on the same audit conditions, with lower pass rates; on the repair task it
reached high scores in 20--45 minutes. The pmover driver task shows the widest
spread: one valid run passed all 30 scenarios in 41 turns and 14 minutes, the
other passed 12 of 30 after 223 turns and 71 minutes, which makes
interoperation the track with the longest horizon.

\subsection{Run validity}
\label{sec:validity}

\input{tables/validity}

Table~\ref{tab:validity} accounts for all 148 launched slots, of which 101
were valid. For sol, the provider's cyber-safety filter stopped 7 of the 44
runs on the xbar audit tasks: none of the 10 runs on revisions 2--3, and 7 of
34 on revision 4 and its variants (netlist 2/7, layout 1/7, obfuscated
4/20). It stopped none of the 29 runs on pmover or pkr. The task
text was identical across slots of a condition, so refusal is stochastic per
run, not a property of the prompt. For a reverse-engineering benchmark this
matters twice: a refusal counted as a failure would misreport capability, and a
refusal rate that varies across conditions is itself a confound. \bench{}
therefore reports refusals separately and draws comparisons only between valid
runs.

\subsection{Case study: a softmax accelerator from its layout}
\label{sec:softmax}

We built an 8-lane int8 softmax accelerator (causal mask, 10-segment
exponential lookup, iterative normalization; about 1.9k gates), ran it through
OpenROAD on ASAP7, and sanitized the layout to L1. Given only the GDSII,
Claude Opus 5.5 extracted a netlist with KLayout, built behavioral cell models,
fixed its own extraction error, probed the design in simulation, identified
max-and-index search, exponential accumulation, lane masking and normalization,
and wrote an RTL that matched the reference on every output in every cycle over
more than 250k simulated cycles, including all 256 lookup differences, in under
ten minutes. Under \bench{} rules this task would be rejected, because a
transcribed netlist also passes an equivalence check. Its value is as a lower
bound: recovering a small datapath from a layout is already cheap, which is why
the benchmark must ask for more than equivalence.

\section{Discussion}

The pilot suggests three design rules that the full release adopts.
\textbf{Ask for what the artifact cannot give.} Equivalence-graded recovery is
solved by transcription (Section~\ref{sec:softmax}), and audit tasks with a
complete specification can be solved without reading the structure
(Section~\ref{sec:how}). Abstract and Interoperate tasks, whose deliverables
need knowledge of parameters or a programming model, carry most of the signal.
\textbf{Measure process, not only outcome.} Stage traces and artifact reads
distinguished strategies that pass rate could not, and they are cheap to
compute from logged trajectories. \textbf{Report validity.} Refusals and
sandbox failures removed a third of the slots; without explicit accounting the
same data would read as a capability gap.

\section{Responsible Scope}
\label{sec:ethics}

Hardware reverse engineering is dual-use. \bench{} uses only original designs
built by the authors on open PDKs. It contains no tasks that defeat IP protection
mechanisms (logic locking, camouflaging, watermark removal), no tasks that
extract secrets (side channels, scan-chain attacks), no tasks against
third-party commercial chips and no tasks that insert malicious logic. Each track
corresponds to legitimate engineering on one's own or licensed hardware: audit,
interoperability, maintenance and repair. Measuring how cheap reverse
engineering has become helps defenders more than attackers, because it tells IP
owners which protections still matter.

\section*{Limitations}

The pilot covers three designs, two agents and 101 valid runs; with 4--10 valid
runs per cell only large effects are detectable, and the null results on
obfuscation are bounded accordingly. Task revisions and provider channels
changed between some batches (noted in Table~\ref{tab:main} and
Figure~\ref{fig:obf}), so we compare only conditions run under the same
revision and channel. Stage labels come from rules over command text and can
misassign multi-purpose scripts; the rules and the extraction scripts are
released for audit. Several tracks (Extract at L3, metal-only ECO, microcode
recovery, parameterized families) are not yet instantiated, and analog layouts
are out of scope. The two agents ran in their native harnesses, so harness and
model effects are confounded. Finally, procedurally generated designs may carry
stylistic regularities that agents could exploit; we will audit this before
release.

\bibliography{refs}

\appendix

\section{Stage Annotation Rules}
\label{app:stages}

Each tool call is reduced to its command text (the shell command for Codex
\texttt{exec\_command} and dsh \texttt{bash}; the JSON arguments otherwise) and
assigned the first matching stage:
\begin{enumerate}
\item \emph{poll} (excluded): \texttt{write\_stdin} without a new command.
\item \emph{author}: \texttt{apply\_patch}, file creation through \texttt{cat >}
  or heredocs, or Python that writes a file.
\item \emph{extract}: KLayout, gdstk/gdspy, \texttt{.gds}/\texttt{.lef}/\texttt{.lyt}
  files, \texttt{read\_liberty}, \texttt{LayoutToNetlist}.
\item \emph{self-validate}: the public checker (\texttt{tbcheck.py check},
  \texttt{pmcheck}, a grade or validate script).
\item \emph{structural analysis}: anonymized net names, the DUT netlist path
  under \texttt{grep}/\texttt{sed}/\texttt{head}/\texttt{cat}, cone and
  fan-in/fan-out queries, ABC, \texttt{read\_verilog}, netlist JSON.
\item \emph{behavioral probing}: simulators (Icarus, Verilator), replay,
  co-simulation, fuzzing, hypothesis and model scripts.
\item \emph{orient}: specification and documentation files, the checker's
  source, and directory listing commands.
\item \emph{other}: everything else (5--20\% of calls).
\end{enumerate}
The rules are coarse by design and are meant for aggregate shares, not for
labeling individual calls; the main ambiguity is Python scripts that both
simulate and write files, which the rules count as authoring. We have not yet
measured agreement with human labels.

\section{Seed Task Details}
\label{app:tasks}

\begin{table}[h]
\centering
\small
\setlength{\tabcolsep}{3pt}
\begin{tabular}{p{1.2cm}p{2.3cm}p{2.85cm}}
\toprule
Task & Given & Graded on \\
\midrule
xbar audit & spec; netlist or L1 GDS + PDK; 3 golden traces & TB passes golden on 3 seeds, $\geq$9/10 coverage points, kills shipped and hidden mutants; 8 isolated reproducers \\
xbar repair & spec; defective netlist (V0/V0r/T2$\times$3/W25); replay tool & per-defect replay vs.\ golden (0--8); pass = 8 \\
pmover driver & L1 GDS + PDK; Verilator co-sim harness; C API & 30 hidden scenarios over init, copy, scatter-gather, checksum and error reporting; pass = 30/30 \\
pkr audit & spec; protected Verilator model with 8 bugs & sign-off TB plus 8 directed tests; 60 hidden checks; pass = 60/60 \\
\bottomrule
\end{tabular}
\caption{Seed tasks used in the pilot.}
\label{tab:tasks}
\end{table}

\end{document}

%% file: tables/main_results.tex
\begin{table*}[t]
\centering
\small
\setlength{\tabcolsep}{4.5pt}
\resizebox{\textwidth}{!}{%
\begin{tabular}{lcccccccc}
\toprule
Condition & Agent & Slots & Pass / valid & Pass rate [95\% CI] & Turns & Minutes & Tokens (M) & Artifact reads \\
\midrule
\multicolumn{9}{l}{\textit{Audit: specification vs.\ implementation (xbar, 8 seeded defects)}} \\
\quad Netlist, task rev.\ 2--3 & sol & 10 & 6/9 & 0.67 [0.35, 0.88] & 42 & 25 & 2.9 & 16 \\
\quad Netlist, task rev.\ 4 & sol & 7 & 3/5 & 0.60 [0.23, 0.88] & 49 & 25 & 4.2 & 5 \\
\quad Sanitized GDS (L1) + PDK & sol & 7 & 2/4 & 0.50 [0.15, 0.85] & 62 & 37 & 5.4 & 10 \\
\addlinespace[2pt]
\multicolumn{9}{l}{\textit{Audit under keyless obfuscation (xbar rev.\ 4)}} \\
\quad V0r re-sanitized & sol & 6 & 2/5 & 0.40 [0.12, 0.77] & 41 & 27 & 3.0 & 6 \\
\quad T2$\times$3 resynthesis & sol & 7 & 3/4 & 0.75 [0.30, 0.95] & 46 & 33 & 3.9 & 7 \\
\quad W25 windowed hiding & sol & 7 & 3/4 & 0.75 [0.30, 0.95] & 54 & 31 & 4.1 & 6 \\
\quad V0 baseline & DS & 6 & 2/5 & 0.40 [0.12, 0.77] & 187 & 81 & 74.0 & 46 \\
\quad V0r re-sanitized & DS & 6 & 0/5 & 0.00 [0.00, 0.43] & 188 & 93 & 64.8 & 46 \\
\quad T2$\times$3 resynthesis & DS & 7 & 1/5 & 0.20 [0.04, 0.62] & 233 & 155 & 85.7 & 32 \\
\quad W25 windowed hiding & DS & 5 & 0/5 & 0.00 [0.00, 0.43] & 270 & 194 & 111.0 & 48 \\
\addlinespace[2pt]
\multicolumn{9}{l}{\textit{Modify: repair the defective netlist (xbar, graded 0--8)}} \\
\quad V0 baseline & DS & 8 & 4/6 & 0.67 [0.30, 0.90] & 117 & 21 & 31.8 & 37 \\
\quad V0r re-sanitized & DS & 14 & 5/8 & 0.62 [0.31, 0.86] & 126 & 44 & 32.5 & 42 \\
\quad T2$\times$3 resynthesis & DS & 14 & 8/10 & 0.80 [0.49, 0.94] & 136 & 44 & 35.9 & 28 \\
\quad W25 windowed hiding & DS & 10 & 7/10 & 0.70 [0.40, 0.89] & 126 & 39 & 30.6 & 31 \\
\addlinespace[2pt]
\multicolumn{9}{l}{\textit{Interoperate: driver for an undocumented DMA peripheral}} \\
\quad Sanitized GDS (L1) + PDK & sol & 13 & 1/2 & 0.50 [0.09, 0.91] & 132 & 43 & 19.3 & 24 \\
\addlinespace[2pt]
\multicolumn{9}{l}{\textit{Black-box reference: audit with a compiled model only (pkr)}} \\
\quad Protected Verilator model & sol & 16 & 5/14 & 0.36 [0.16, 0.61] & 82 & 40 & 8.8 & 0 \\
\bottomrule
\end{tabular}}
\caption{Pilot rollouts. \emph{Slots} counts every launched run; \emph{valid} excludes runs that ended in an
infrastructure error or a provider safety refusal (Table~\ref{tab:validity}). Pass rates carry Wilson 95\% intervals.
Turns, minutes, input tokens and artifact reads (tool calls that open the DUT netlist or layout) are medians over
valid runs. sol = GPT-6.1-sol (xhigh, Codex harness); DS = DeepSeek-flash (max effort, DeepSeek harness).
For the repair task, pass means all 8 defects repaired; Figure~\ref{fig:obf} shows the full 0--8 scores.
The sol rev.\ 4 netlist row (V0) ran on a different provider channel from V0r/T2$\times$3/W25, so V0r is the controlled baseline for sol.}
\label{tab:main}
\end{table*}

%% file: tables/validity.tex
\begin{table}[t]
\centering
\small
\setlength{\tabcolsep}{4pt}
\resizebox{\columnwidth}{!}{%
\begin{tabular}{lccccc}
\toprule
Agent & Launched & Valid & Refusal & Agent/API & Sandbox \\
\midrule
GPT-6.1-sol & 78 & 47 & 7 & 14 & 10 \\
DeepSeek-flash & 70 & 54 & 0 & 5 & 11 \\
\bottomrule
\end{tabular}}
\caption{Run accounting for all pilot slots. \emph{Refusal}: the provider's cyber-safety filter stopped the run
(\texttt{AgentSafetyRefusalError}); \emph{Agent/API}: agent process crash or provider overload;
\emph{Sandbox}: cold-boot or connection failures of the execution sandbox. Invalid runs are excluded from pass
rates and are never re-graded as failures.}
\label{tab:validity}
\end{table}